\documentclass[lettersize,journal]{IEEEtran}
\usepackage{amsmath,amsfonts}
\usepackage{algorithmic}
\usepackage{algorithm}
\usepackage{array}
\usepackage[caption=false,font=normalsize,labelfont=sf,textfont=sf]{subfig}
\usepackage{textcomp}
\usepackage{stfloats}
\usepackage{url}
\usepackage{verbatim}
\usepackage{graphicx}
\usepackage{cite}
\usepackage{booktabs}
\usepackage{pifont}
\usepackage{multirow}
\usepackage[hidelinks]{hyperref}
\usepackage{enumitem}
\usepackage[table]{xcolor}
\definecolor{lightgreen}{RGB}{226,239,218}
\usepackage{float}
\newcommand{\yes}{\ding{51}}
\newcommand{\no}{\ding{55}}
\begin{document}


\title{MERIT: Experience-Guided Memory for Long-Horizon Persona-Specific Persuasive Dialogue}

\author{Rohan Kirti, Akash Ghosh, Pratik Amrit, Aryan Vats, Niladri Ghosh, and Sriparna Saha%
\thanks{Rohan Kirti, Akash Ghosh, Pratik Amrit, and Sriparna Saha are with the Department of Computer Science and Engineering, Indian Institute of Technology Patna, India (e-mail: \href{mailto:rohan_2421cs09@iitp.ac.in}{rohan\_2421cs09@iitp.ac.in}).}%
\thanks{Aryan Vats is with the Department of Mathematics and Computing, Indian Institute of Technology Patna, India.}%
\thanks{Niladri Ghosh is with the Department of Computer Science, Ramakrishna Mission Vivekananda Educational and Research Institute, India.}%
}


\maketitle

\begin{abstract}
Effective long-horizon persuasion in recurring goal-oriented settings requires more than factual memory: it must track how a user's trust, negotiation behavior, and strategic receptiveness evolve across interactions. Existing persuasive dialogue systems are largely single-session, static, and unimodal, while cross-session memory systems, though temporally aware, retain only factual content and overlook persuasion-specific dynamics. We propose \textbf{\textit{MERIT}}
(\underline{M}emory-augmented \underline{E}xperience-\underline{R}ationale
for \underline{I}terative \underline{T}emporal persuasion), which pairs a graph-based memory
of four persuasion-specific memory types with a per-persona experience rationale library
that records which strategies succeed for each user and why. Together they enable adaptive persuasion through prompt composition alone, with no parameter updates. To support this setting, we introduce \textbf{\textit{PERSIST }}(\textbf{PERS}ona-driven \textbf{I}n\textbf{\textit{S}}urance Interactions over \textbf{T}ime), a dataset of 60 curated personas spanning 10+ years of motor insurance renewal interactions, with probabilistic life events that shift negotiation behavior and receptiveness across sessions, verified through strict human-in-the-loop annotation. \textbf{\textit{MERIT}} surpasses memory and experience-based baselines on every metric across two policy backbones, reaching persuasive-appeal scores of 4.10 and 4.74 (out of 5).
\end{abstract}


\section{Introduction}

Long-term conversational interactions are increasingly common as Large Language Models (LLMs) and Multimodal Language Models (MLLMs) are deployed in applications requiring sustained user engagement \cite{xi2025rise, wang2024survey,ghosh2404exploring,ghosh2026carepilot}. In settings such as insurance sales, healthcare users return across sessions spanning months or years, requiring agents to remember prior conversations and adapt to evolving user behavior, negotiation patterns, and trust dynamics. Users also share images to ground their concerns in visual contexts that cannot be fully captured through language alone\cite{ghosh2024clipsyntel,ghosh-etal-2024-sights}. These scenarios require agents to maintain persistent, persona-specific understanding across time and modalities rather than treating each interaction independently.

Recent advances in computational persuasion show that LLMs and MLLMs can generate personalized persuasive strategies across domains such as sales, negotiation, recommendation, and adversarial dialogue \cite{bozdag2025must, rogiers2024persuasion, kirti2026can}. Prior work has also explored persona-aware and context-adaptive persuasive systems for personalized conversational agents \cite{tiwari2022persona, tiwari2023towards}.

However, existing persuasive dialogue systems primarily operate in isolated single-session settings or unimodal setups, relying on static persona representations that do not evolve through continued interaction \cite{samad2022empathetic, chen2022seamlessly, tseng2024two, zhang2018personalizing, li2016persona}. Although memory-augmented architectures preserve conversational context across sessions, they focus primarily on factual retention and consistency rather than persuasion dynamics, such as strategy effectiveness, negotiation behavior, and trust evolution \cite{zhong2024memorybank, wang2025recursively, huang2026mem}. Crucially, no existing framework captures accumulated strategic experience, that is, \textit{structured knowledge of which persuasive strategies proved effective for a specific user over time and why, leaving agents unable to transfer lessons from past persuasive successes and failures into future interactions.}

\textbf{Present Work:} To address this gap, a natural approach to long-horizon persuasive dialogue is to equip the models with external memory of prior interactions through retrieved session summaries, user profiles, or past negotiation outcomes. While such memory helps recall recurring objections, prior decisions, and evolving conversational context, remembering past interactions is fundamentally different from knowing how to persuade effectively in the current interaction. 
This calls for two complementary capabilities: \textit{a memory structure that preserves persuasion-specific dynamics across years, and an explicit mechanism for accumulating strategic experience like what works, for whom, and why, across repeated interactions with the same user.}

To support this setting, we introduce \textbf{\textit{PERSIST}}, a multimodal persona-driven insurance dialogue dataset capturing longitudinal persuasive interactions across 60 personas over multiple years of recurring conversations. 
Building upon this, we propose \textbf{\textit{MERIT}}, a memory-augmented framework for iterative temporal persuasion that decomposes long-horizon persuasive adaptation into two complementary components. \textit{First, a graph-based memory architecture organizes interaction history into four specialized representations: episodic session context, strategy effectiveness, latent persuasion states, and negotiation concession patterns.} These representations are connected through a temporal graph that supports semantically targeted retrieval. \textit{Second, MERIT introduces persona-specific experience rationales constructed through referee-evaluated response comparisons, capturing not only which strategies succeed for a given user but also why they are effective.} These accumulated rationales form a per-persona experience library that enables adaptive strategic persuasion through prompt composition rather than parameter updates.


\textbf{Contributions:} Our key contributions are summarised below:
\begin{enumerate}[label=\textbf{(\roman*)}, leftmargin=*]
    \item \textbf{Task Formulation.} We formalize the \textit{first treatment of persuasion as a longitudinal, multimodal, persona-specific process}, where strategies must adapt across multi-year recurring interactions with the same user, establishing it as a long-horizon, persona-specific process.
    \item \textbf{Dataset.} We introduce \textbf{\textit{PERSIST}}, the first multimodal longitudinal persuasive dialogue dataset spanning 60 personas over 10+ years of motor insurance interactions, capturing evolving user behavior, negotiation patterns, and trust dynamics across recurring sessions.
    \item \textbf{Framework.} We propose \textbf{\textit{MERIT}}, which combines a frozen policy model with graph-based memory comprising four persuasion-aware memory types and persona-specific experience rationales that capture effective strategies and their underlying reasons, enabling longitudinal personalization without parameter updates.
    \item \textbf{Evaluation.} We conduct extensive quantitative and qualitative evaluations on \textit{PERSIST}, demonstrating that \textbf{\textit{MERIT}} effectively preserves long-term persuasion dynamics and enables persona-specific persuasive response generation across multi-year interactions.
\end{enumerate}

\section{Related Works}

\subsection{Long-Horizon Memory and Experience-Based Reasoning}

Memory-augmented architectures for LLM agents have evolved from flat retrieval to structured representations. \cite{zhong2024memorybank} introduced forgetting-curve-based memory management, \cite{wang2025recursively} proposed recursive summarization for long-term dialogue consistency, and \cite{packer2023memgpt} applied OS-inspired hierarchical paging. More recent systems pursue richer structure: \cite{chhikara2025mem0} abstracts memory as an entity-centric graph layer, \cite{xu2026mem} builds self-evolving knowledge networks via Zettelkasten-inspired linking, \cite{huang2026mem} models heterogeneous user history with persona-aware updates, \cite{tan2025prospect} refines retrieval online through topic-based memory organization and citation-driven reranking, \cite{yan2026adamem} organizes history into different memories with question-conditioned retrieval, and \cite{yuan2025personalized} introduces evolving conditional memory for personalized LLM assistants.

A complementary line of work addresses strategic reasoning through experience-based learning without gradient updates. \cite{madaan2023self} showed that LLMs can improve outputs through iterative self-feedback within a single episode, while \cite{shinn2023reflexion} extended this to cross-episode verbal reinforcement. \cite{zhao2024expel} generalized the paradigm by extracting reusable insights from success-failure trajectory pairs, and \cite{kim2025principles} derived synthetic strategy memories through offline self-play for proactive dialogue. \looseness=-1\noindent\underline{\smash{\textit{\textbf{Research Gap.}}}} \textit{Collectively, these works leaves open how persuasion-specific memory and persona-specific strategy adaptation can be modeled across longitudinal interactions in multimodal long horizon setup.}

\begin{table}[!t]
\caption{PERSIST statistics (left) and comparison with existing
benchmarks (right). MM: multimodal; LH: long-horizon;
PD: persona-driven; ED: evolving dynamics.}
\label{tab:dataset}
\centering
\scriptsize
\begin{minipage}[t]{0.42\columnwidth}
\centering
\begin{tabular*}{\linewidth}[t]{@{}l@{\extracolsep{\fill}}r@{}}
\toprule
\textbf{Statistic} & \textbf{Value} \\
\midrule
\multicolumn{2}{@{}l}{\textit{Scale}} \\
Personas                  & 60 \\
Dialogues                 & 643 \\
Utterances                & 20{,}142 \\
Unique images             & 292 \\
\midrule
\multicolumn{2}{@{}l}{\textit{Design}} \\
Persuasion strategies     & 6 \\
User intents              & 8 \\
Life-event types          & 10 \\
\midrule
\multicolumn{2}{@{}l}{\textit{Longitudinal}} \\
Time span                 & 2015--2025 \\
Avg.\ dialogues/persona   & 10.88 \\
Avg.\ utterances/dialogue & 31.3 \\
\bottomrule
\end{tabular*}
\end{minipage}%
\hfill
\begin{minipage}[t]{0.55\columnwidth}
\centering
\setlength{\tabcolsep}{2.5pt}
\begin{tabular*}{\linewidth}[t]{@{}l@{\extracolsep{\fill}}cccc@{}}
\toprule
\textbf{Dataset} & \textbf{MM} & \textbf{LH} & \textbf{PD} & \textbf{ED} \\
\midrule
\multicolumn{5}{@{}l}{\textit{Negotiation datasets}} \\
DealOrNoDeal \cite{lewis2017deal}          & \no  & \no  & \no  & \no \\
Craigslist Bargain \cite{he2018decoupling} & \no  & \no  & \no  & \no \\
CaSiNo \cite{chawla2021casino}             & \no  & \no  & \no  & \no \\
DEAL \cite{priya2024trip}                  & \no  & \no  & \yes & \no \\
PACT \cite{priya2025we}                    & \no  & \no  & \yes & \no \\
\midrule
\multicolumn{5}{@{}l}{\textit{Persuasion datasets}} \\
PFG \cite{wang2019persuasion}              & \no  & \no  & \no  & \no \\
PVP \cite{kim2025pvp}                      & \yes & \no  & \no  & \no \\
MMPersuade \cite{qiu2025mmpersuade}        & \yes & \no  & \no  & \no \\
PPMD \cite{tiwari2022persona}              & \yes & \no  & \yes & \no \\
Talk2AI \cite{carrillo2026talk2ai}         & \no  & \no  & \no  & \no \\
\midrule
\textbf{PERSIST (Ours)} & \textbf{\yes} & \textbf{\yes}
                        & \textbf{\yes} & \textbf{\yes} \\
\bottomrule
\end{tabular*}
\end{minipage}
\end{table}

\subsection{Multi-Session Dialogue and Persuasion}

Multi-session dialogue research has progressively advanced long-term conversational consistency. \cite{xu2022beyond} extended persona-based dialogue to multi-session settings, while \cite{maharana2024evaluating} introduced longitudinal multimodal interactions across extended periods. \cite{kim2025share} incorporated shared memory across sessions, and \cite{bae2022keep} studied memory management through selective updating and forgetting mechanisms. More recently, \cite{li2025hello} proposed personalized agents with evolving user models for sustained interactions. Persuasive dialogue research has largely focused on single-session interactions, including strategy-aware persuasion \cite{wang2019persuasion}, goal-oriented negotiation and concession strategies \cite{lewis2017deal,chawla2021casino}, and later work on strategy-aware generation, empathetic persuasion, and adaptive persuasion, and reinforcement learning-based expert selection for persuasion \cite{he2018decoupling,samad2022empathetic,chen2022seamlessly,kirti2026persuarl}. \looseness=-1\noindent\underline{\smash{\textit{\textbf{Research Gap.}}}}\textit{ However, existing negotiation datasets such as DealOrNoDeal and CaSiNo do not model persona evolution, life events, multimodal context, or multi-year interaction histories. In contrast, PERSIST captures 10+ years of interactions per persona, where negotiation behavior evolves through probabilistic life events and is grounded in multimodal vehicle context, making long-term user-specific persuasion dynamics directly observable.}


\section{Methodology}
We then introduce \textbf{\textit{MERIT}} (Sec. \ref{sec:Merit}),
a framework for iterative temporal persuasion that combines a
graph-based memory with persona-specific experience rationales
capturing which strategies prove effective for each user over time,
with no parameter updates to the policy model.

\subsection{PERSIST Dataset}
\label{sec:Persist}
\noindent\textbf{Overview.} We introduce PERSIST, a multimodal persona-driven dataset of longitudinal persuasive dialogues in the motor insurance domain. Each persona has at least one insurance renewal conversation per year over a span of 10+ years, forming a connected interaction history rather than a set of independent dialogues. Before selected sessions, the persona may undergo a probabilistic life event, such as an accident, an income change, or a major purchase, that affects its negotiation-relevant attributes. \textit{The life-event sampling process and its effects on persona attributes are described in Appendix~\ref{sec:lifeevents}}. Each session is also grounded in a vehicle image, which is processed by the Gemma 3 4B model to anchor the conversation in a concrete multimodal context. As summarized in Table~\ref{tab:dataset}, PERSIST is the only existing dataset that is simultaneously multimodal, long-horizon, persona-driven, and captures evolving negotiation dynamics. PERSIST adopts a human-in-the-loop approach, combining LLM-generated dialogues with human-authored turns, seed utterances, and quality validation, following established practices in synthetic dialogue dataset construction \cite{maharana2024evaluating,jang2023conversation,kim2023soda}.

\paragraph{\textbf{Persona Construction}} Each persona is a discrete combination of a base customer profile, a dominant Big Five trait \cite{digman1990personality}, and an income bracket. The Cartesian product of these three axes yields $7 \times 5 \times 3 = 105$ candidate configurations. Three domain experts independently reviewed these candidates and, using majority voting, curated a final set of 60 personas based on behavioral distinctiveness and contextual plausibility. The resulting set covers all seven base profiles, five dominant Big Five traits, and three income brackets, with approximately balanced representation: 8--9 personas per base profile, 12 per dominant trait, and 20 per income bracket. Each dominant trait is represented across all three income brackets. The persona configuration space of \textbf{\textit{PERSIST}} is shown in Table \ref{tab:persona-axes}.

\begin{table}[t]
\caption{The three-axis persona configuration space of \textbf{\textit{PERSIST}}.}
\label{tab:persona-axes}
\centering
\small
\begin{tabular}{@{}lp{5.2cm}@{}}
\toprule
\textbf{Dimension} & \textbf{Values} \\
\midrule
Base Profile (7)       & enthusiastic, tech-savvy, skeptical, detail-oriented, cautious, first-time buyer, budget-conscious \\
\midrule
Big Five Dominant (5)  & openness, conscientiousness, extraversion, agreeableness, neuroticism \\
\midrule
Income Bracket (3)     & low, mid, high \\
\bottomrule
\end{tabular}
\end{table}

\paragraph{\textbf{Dialogue Generation}} Each conversation is generated turn by turn by LLMs through a multi-agent pipeline coordinated by a dialogue orchestrator, with human-authored turns incorporated for selected scenarios and human oversight applied during dataset validation. When a vehicle image is provided, Gemma 3 4B IT extracts a structured textual description that is appended to the conversation context. Given the current user utterance and prior context, the orchestrator activates relevant expert modules: Sentiment, Motive, Keyterm, and Engagement Experts, which analyze the user's emotional state, communicative intent, salient insurance terms, and persuasion-related signals, respectively. Each expert returns an analysis with an explicit rationale and may consult a shared motor-insurance knowledge base containing insurer information, premium pricing, depreciation schedules, coverage details, and policy terms. The orchestrator aggregates the expert outputs and passes them to the \textit{Persuasion Synthesizer}, which selects a persuasion strategy and generates the agent response.

Dialogue generation uses eight user intents and six agent persuasion strategies. The user intents capture communicative goals ranging from session onset, coverage inquiry, and quote solicitation to concern disclosure, clarification request, premium negotiation, renewal commitment, and offer decline. The six agent persuasion strategies are adopted from the persuasive sales-dialogue taxonomy of \cite{tiwari2022persona}: \emph{Default}, which provides neutral information without an explicit persuasive appeal; \emph{Credibility}, emphasizing insurer reliability; \emph{Emotional}, addressing the user's feelings and concerns; \emph{Logical}, using facts and reasoning; \emph{Personal}, drawing on recommendations and social proof; and \emph{Persona-Adaptive}, tailoring the appeal to the user's attributes and priorities. At each turn, the agent strategy is selected before the user's next intent is chosen based on the agent's latest response and prior conversation context.

Strategy selection is governed by a per-persona complete directed graph over the six persuasion strategies, including self-loops. Its weight matrix $W_p$ encodes the propensity to select strategy $s'$ after strategy $s$, with $W_p(s,s')$ denoting the corresponding transition weight. The matrix is initialized from a shared prior and updated over the persona's training-set conversations. At turn $t$, the \emph{Spectrum} mechanism adapts these long-term transition preferences to the current context. Conditioned on the user's current intent, inferred sentiment, and persona profile, Gemini 2.5 Pro \cite{comanici2025gemini} assigns each candidate strategy $s'\in\mathcal{S}$ a positive context-fit multiplier $m_t(s')$, where values above, below, or near $1$ indicate a good, poor, or neutral fit, respectively. These multipliers reshape the outgoing habit weights, which are normalized into a turn-specific distribution:
\begin{equation}
\pi_t(s') =
\frac{W_p(z_{t-1},s')\,m_t(s')}
{\sum_{s''\in\mathcal{S}}W_p(z_{t-1},s'')\,m_t(s'')}.
\end{equation}
The next strategy is sampled as $z_t\sim\mathrm{Categorical}(\pi_t)$, preserving strategic diversity while favoring strategies suited to the current interaction. Spectrum is recomputed at every turn and does not modify the underlying habit weights.

After the agent response and simulated user reply are generated, a Gemini 2.5 Pro judge evaluates the exchange. The mean quality score $q_t\in[0,1]$ over the five most recent turns determines an additive update to only the traversed edge:
\begin{equation}
W_p(z_{t-1},z_t)\leftarrow
\max\!\left(0.01,\,
W_p(z_{t-1},z_t)+\delta(q_t)\right),
\end{equation}
where
\begin{equation}
\delta(q_t)=
\begin{cases}
+0.0050, & q_t>0.80,\\
+0.0025, & 0.60<q_t\leq0.80,\\
-0.0025, & 0.40\leq q_t\leq0.60,\\
-0.0050, & q_t<0.40.
\end{cases}
\end{equation}
These small updates reinforce or weaken strategy transitions according to recent interaction quality, allowing persona-specific strategic tendencies to evolve over time. Unlike Spectrum's temporary context adjustment, the habit update changes the underlying transition weights. Finally, conditioned on the agent's response and conversation history, Gemini 2.5 Pro selects the most fitting intent from the eight-intent taxonomy and generates a persona-consistent user reply.

\begin{figure*}[!t]
\centering
\includegraphics[width=\textwidth]{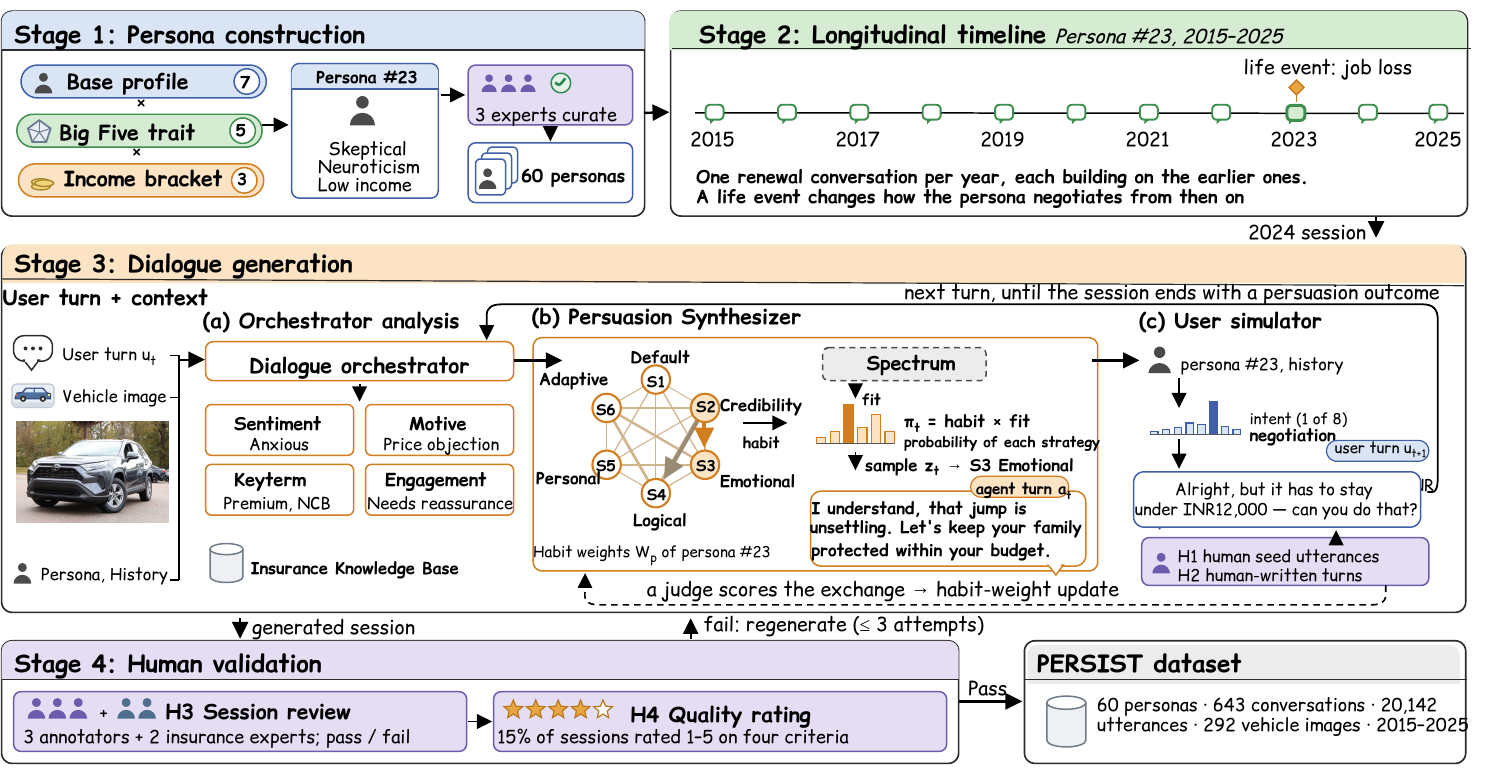}
\caption{PERSIST construction: (1) 60 personas defined by profile, Big Five trait, and income; (2) annual renewal dialogues (2015--2025) shaped by life events; (3) multi-agent dialogue generation with persona-specific persuasion strategies and human inputs (H1, H2); and (4) expert review (H3) and sample quality assessment (H4). Purple indicates human input.}
\label{fig:diagram}
\end{figure*}

\paragraph{\textbf{Human-Grounded Dataset Construction}}
\textit{PERSIST} incorporates human input at four stages:

\textit{Seed Data.}
Human annotators formulate short customer utterances for predefined motor-insurance scenarios, covering intents such as Concern Disclosure, Clarification Request, Premium Negotiation etc. Each utterance is tagged with its scenario, intended user intent, and relevant persona context. These utterances form a seed bank used as in-context examples to guide the user simulator in generating diverse, persona-consistent customer responses. The seed utterances are used solely as generation references and are not included in the released dataset.

\textit{Human-Authored User Turns.}
For selected scenarios, human annotators formulate customer turns that are incorporated into the released dialogues. To maintain longitudinal consistency, annotators receive the persona profile, summaries of previous sessions, and relevant insurance and policy history. The associated vehicle image and its description are also provided when applicable. The LLM generates the corresponding agent responses and remaining turns, conditioned on the human-scripted turn, persona profile, and interaction history. Original human-scripted turns are retained to verify that their intent and content are preserved during integration. Provenance metadata distinguishes human-scripted, model-generated, and human-edited turns.

\textit{Validation.}
During dataset construction, each generated session is reviewed by one of three postgraduate annotators. Additionally, 96 sessions (15\% of the dataset) are independently reviewed by all three annotators, with disagreements resolved by majority vote. Reviewers check life-event plausibility, chronological consistency, and consistency between previous interactions and subsequent insurance decisions, marking each criterion as pass or fail. For sessions containing human-authored turns, they also verify that the original intent and content are preserved. Two domain experts examine insurance-specific correctness, including policy-related statements and coverage descriptions. Sessions that fail these checks are corrected or regenerated using reviewer feedback, with a maximum of three generation attempts, including the initial attempt. Sessions that still fail after three attempts are discarded.

\textit{Independent Human Evaluation of Dialogue Quality.}
To further assess dialogue quality, 15\% of the generated dialogues are independently evaluated, covering different personas, time periods, and negotiation scenarios. Annotators assess persona consistency, temporal consistency, intent alignment, and persuasion-strategy
appropriateness using a five-point Likert scale (1 = very poor, 2 = poor, 3 = adequate, 4 = good, and 5 = excellent). Insurance-specific factual correctness is also examined. \textit{Inter-annotator agreements are reported in Appendix~\ref{sec:IAA}.}

\begin{figure*}[t]
\centering
\includegraphics[width=\textwidth]{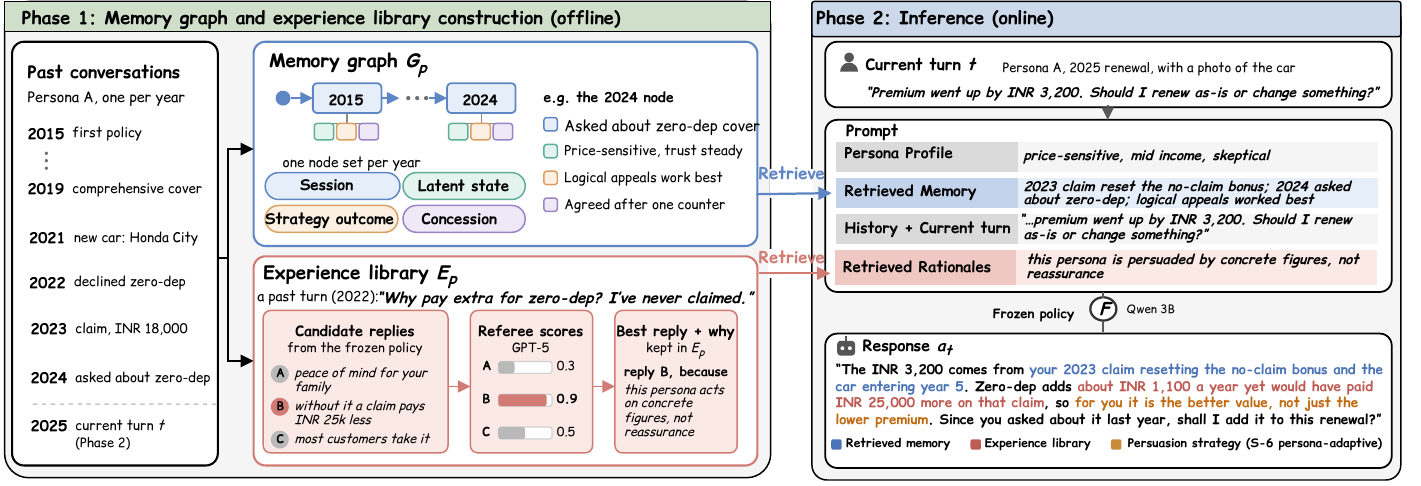}

\caption{\textbf{\textit{MERIT}} experience library construction and inference. \textbf{ Library Construction (offline):} for each turn, MERIT retrieves persona-specific memory, generates 3 candidate rollouts, and a referee scores them and stores the best in the per-persona experience library. \textbf{Inference (Online):} MERIT retrieves persona-specific memory and the experience rationales, composes an augmented prompt, and the frozen policy model generates the response.}
\label{fig:architecture}
\end{figure*}

\subsection{MERIT}
\label{sec:Merit}
\textbf{Problem Formulation.} We study \textit{longitudinal persona-specific persuasive dialogue}, a setting in which the agent must adapt persuasion strategies across multi-year recurring interactions with the same user, accumulating and transferring strategic experience across sessions.
Let $\mathcal{P}$ denote the set of personas. For a persona 
$p \in \mathcal{P}$, the interaction history is an ordered 
sequence of conversations across years,
\begin{equation}
    C_p = \langle c_p^1, c_p^2, \ldots, c_p^{N_p} \rangle,
\end{equation}
where each $c_p^i$ is one conversation ordered in time. Each
conversation is a sequence of turns,
$c_p^i = \langle (u_1, a_1, d_1), \ldots,
(u_T, a_T, d_T) \rangle$,
where $u_t$ is the user utterance, $a_t$ is the agent
response, and $d_t$ is an optional textual description of the
user's vehicle. When the user shares a vehicle photograph $v_t$
at turn $t$, $d_t$ is the structured description extracted from
$v_t$ by Gemma~3~4B \cite{gemma_2025};
otherwise $d_t = \emptyset$. Each conversation also yields a
persuasion outcome in the form $o_p^i \in \{0, 1\}$, indicating whether
the agent successfully persuaded persona $p$ in year $i$.

Each conversation further yields four persona-memory 
components, namely session memory, latent persuasion state, 
strategy-outcome memory, and concession memory, collectively 
written $m_p^i$. The memory graph $\mathcal{G}_p$ organizes 
$\{m_p^1, \ldots, m_p^{N_p}\}$ across years. A 
persona-specific experience library $\mathcal{E}_p$ 
accumulates strategic knowledge of which persuasive choices 
succeed for the persona over time and, importantly, 
\textit{why} they are effective.

\textbf{Task.} Given a persona $p$, a current conversation 
context $h_t$, the current user turn $(u_t, d_t)$,
the memory graph $\mathcal{G}_p$, and the experience library 
$\mathcal{E}_p$, the goal is to generate the agent response

\begin{equation}
\label{eq:task}
    a_t = \mathcal{F}\big(u_t,\; d_t,\; h_t,\;
    R(\mathcal{G}_p, u_t, h_t),\; R_E(\mathcal{E}_p, u_t)\big),
\end{equation}
where $R(\cdot)$ retrieves the relevant memory subset for
the current context, $R_E(\cdot)$ retrieves the Top-$K$ experience rationales most similar to the current query, and $\mathcal{F}$ is a \textbf{frozen} policy model conditioned through prompt composition. No parameters of $\mathcal{F}$ are updated at any point, and adaptation arises entirely from the constructed memory graph and experience library.

\subsubsection{\textbf{Memory Construction}}
\label{sec:memory-construction}
Long-horizon persuasion is heterogeneous: it depends not 
only on what was discussed in past conversations, but also 
on a persona's current trust and receptiveness, which 
strategies have proven effective, and how they negotiate. 
Prior memory-augmented systems~\cite{zhong2024memorybank, wang2025recursively} address temporal consistency through 
flat retrieval of session summaries, but a single unified 
representation cannot simultaneously capture episodic 
content, evolving psychological states, strategy 
effectiveness, and negotiation dynamics. We therefore organize each persona's history into four specialized memory components, each designed to capture a distinct facet of persuasion that the others cannot (Figure \ref{fig:architecture}). Formally, for a persona $p$ at year $i$, the memory is defined as
\[
m_p^i = \left\{
M^{\text{sess}}_{p,i},\;
M^{\text{lat}}_{p,i},\;
M^{\text{strat}}_{p,i},\;
M^{\text{conc}}_{p,i}
\right\},
\]
where $M^{\text{sess}}_{p,i}$ denotes session memory, $M^{\text{lat}}_{p,i}$ captures the latent persuasion state, $M^{\text{strat}}_{p,i}$ stores strategy--outcome information, and $M^{\text{conc}}_{p,i}$ represents concession memory.

\paragraph{\textbf{Session Memory}}
The session memory $M^{\text{sess}}_{p,i}$ records the episodic content of a single conversation. Given a conversation $c_p^i$, an LLM produces a natural-language summary of the interaction, $\sigma_p^i = \text{LLM}(c_p^i)$. This summary is stored together with the persona profile $\pi_p$, comprising the income bracket, dominant Big Five trait, and base personality type, and the final outcome $o_p^i$ of the conversation: $M^{\text{sess}}_{p,i} = \{\, \sigma_p^i,\; \pi_p,\; o_p^i \,\}$. Session memory thus serves as the episodic anchor of the persona's history, recording what happened in each individual year.

\paragraph{\textbf{Latent Persuasion Memory}} The latent persuasion state $M^{\text{lat}}_{p,i}$ gives a 
compact, continuous view of how persuadable a persona is. 
Unlike prior systems that track only factual user 
preferences~\cite{huang2026mem, yuan2025personalized}, we 
model  psychologically grounded 
variables: trust, price 
sensitivity, decision readiness, commitment, strategy 
responsiveness. Trust and price sensitivity are maintained as smoothed running values via an exponential moving average, $x_t = \gamma \, x_{t-1} + (1 - \gamma)\, x^{\text{sig}}_t$, so the persona's disposition shifts gradually rather than abruptly, where $x^{\text{sig}}_t$ is the per-turn signal and $\gamma$ a decay factor. Decision readiness, reflecting how close the persona is to committing, is estimated by an LLM at each turn, and commitment is its average over the conversation. Strategy responsiveness captures how strongly logical, emotional, and credibility-based persuasion have each shifted the persona's readiness.

\paragraph{\textbf{Strategy-Outcome Memory}}
While prior experience-based methods~\cite{zhao2024expel, shinn2023reflexion}
distill strategy knowledge at the episode level, we maintain a continuous
per-turn weighted record that captures fine-grained strategy effectiveness as
persuasion unfolds within a conversation. The strategy-outcome memory
$M^{\text{strat}}_{p,i}$ records how effective each persuasion strategy has
been for the persona, as a weighted set over strategy types
$s \in \mathcal{S}$ (logical, emotional, credibility-based, and others). After
each turn, the persona's mean decision readiness over the most recent turns is
mapped through a banded scheme to an update step $\Delta$, and the weight of
the strategy used in that turn is updated cumulatively,
$w_s \leftarrow w_s + \alpha\,\Delta$, where $\alpha$ is a learning rate.
Higher recent readiness produces larger positive weight changes, so strategies
used while the persona is progressing toward a decision accumulate higher
weight. This memory is maintained per year and serves purely as a record of
strategy effectiveness for retrieval, distinct from any policy used to select
strategies during generation.

\paragraph{\textbf{Concession Memory}} 
Negotiation and concession behavior are central to 
insurance persuasion yet entirely absent from existing 
memory architectures~\cite{zhong2024memorybank, 
wang2025recursively}. The concession memory $M^{\text{conc}}_{p,i}$ captures the negotiation behavior in a conversation. It pairs a logged record $\nu_p^i$ of the negotiation, including the initial price, the user's discount requests, the agent's offers, the final counter, and the outcome, with an LLM-generated summary describing how the negotiation unfolded and which concessions proved effective: $M^{\text{conc}}_{p,i} = \{\, \nu_p^i,\; \text{LLM}(c_p^i) \,\}$. 

\noindent\textbf{Memory Graph and Retrieval.}
$\mathcal{G}_p$ contains one node per memory component per year, a persona-profile root, and three aggregate nodes (latent, strategy, concession) storing information from prior years. Edges connect nodes of the same type across consecutive years, all four nodes within each year, and last-year nodes to their aggregates. $R(\cdot)$ always returns the three aggregates, ranks session nodes using $\lambda\cos(e_q,e_s)+(1-\lambda)\,\mathrm{BM25}(q,s)$ ($\lambda=0.8$, EmbeddingGemma~\cite{vera2025embeddinggemma} embeddings, $q$ = current utterance plus three preceding turns), expands the top two sessions to neighboring sessions in the same or next year. Only prior years are accessible.

\subsubsection{\textbf{Experience Rationale}}

The memory captures prior interactions and persuasion dynamics, but not how the agent should act in future interactions. To address this, MERIT constructs a per-persona experience rationale library $\mathcal{E}_p$ offline using the earliest 80\% of each persona's conversations. For each turn in this subset, the frozen policy model generates three candidate responses, which a referee model (GPT-5 \cite{openai2025gpt5}) scores for persuasive quality and uses to produce a rationale explaining the strongest response's effectiveness. The selected response and its rationale are stored in $\mathcal{E}_p$ along with relevant metadata. During inference, MERIT retrieves the Top-$K$ ($K=8$) rationales most semantically similar to the current query and incorporates them into the prompt alongside the retrieved memory and conversation history. This enables the policy model to reuse persona-specific persuasive experience without parameter updates.

\section{Experiments}
We evaluate \textbf{\textit{PERSIST}} and \textbf{\textit{MERIT}} through experiments designed to answer four research questions:
\textbf{RQ1)} Does MERIT generate more effective persuasive responses than memory and experience-based baselines?
\textbf{RQ2)}  Does MERIT preserve long-horizon persuasion dynamics across multi-year interactions? 
\textbf{RQ3)} Does the per-persona experience rationale library improve persuasion beyond memory retrieval alone?
\textbf{RQ4)} Are MERIT's gains robust across policy backbones differing in family and scale?

\begin{table*}[t]
\caption{Automatic evaluation results of \textbf{\textit{MERIT}} with two policy backbones: \textit{Qwen 2.5 3B Instruct}~\cite{qwen2.5} and GLM-5.1~\cite{zeng2026glm}. On both backbones, \textbf{\textit{MERIT}} outperforms all baselines. All improvements are statistically significant at the 5\% significance level based on a t-test.}
\label{tab:main_results}
\centering
\footnotesize
\begin{tabular}{l|l cccccc}
\toprule
& \textbf{Method} & \textbf{BLEU-2}$\uparrow$ & \textbf{BERTScore}$\uparrow$ & \textbf{ROUGE-1}$\uparrow$ & \textbf{METEOR}$\uparrow$ & \textbf{PAS}$\uparrow$ & \textbf{CAS}$\uparrow$ \\
\midrule
\multicolumn{8}{c}{\textbf{Backbone: Qwen 2.5 3B Instruct}} \\
\midrule
\multirow{11}{*}{\textit{\textbf{Baselines}}}
 & Vanilla LLM                                & 0.056 & 0.593 & 0.327 & 0.094 & 3.245 & 2.572 \\
 & Sliding Window                             & 0.071 & 0.592 & 0.355 & 0.101 & 3.497 & 2.914 \\
 & Conversation Memory                        & 0.059 & 0.597 & 0.334 & 0.095 & 3.147 & 2.767 \\
 \cmidrule{2-8}
 & Session-level RAG                          & 0.063 & 0.598 & 0.341 & 0.097 & 3.785 & 2.826 \\
 & Turn-level RAG                             & 0.058 & 0.603 & 0.337 & 0.095 & 3.645 & 2.869 \\
 \cmidrule{2-8}
 & MemoryBank~\cite{zhong2024memorybank}      & 0.048 & 0.540 & 0.273 & 0.087 & 3.660 & 1.991 \\
 & MemPal~\cite{huang2026mem}                 & 0.099 & 0.585 & 0.334 & 0.086 & 2.811 & 2.740 \\
 & ConditionMem~\cite{yuan2025personalized}   & 0.044 & 0.577 & 0.288 & 0.079 & 2.450 & 2.288 \\
 & RecurSum~\cite{wang2025recursively}        & 0.100 & 0.600 & 0.380 & 0.110 & 2.930 & 3.434 \\
 \cmidrule{2-8}
 & Reflexion~\cite{shinn2023reflexion}        & 0.061 & 0.591 & 0.322 & 0.181 & 2.828 & 2.280 \\
 & ExpeL~\cite{zhao2024expel}                 & 0.074 & 0.586 & 0.325 & 0.200 & 2.794 & 1.843 \\
\midrule
\rowcolor{lightgreen}
\textit{\textbf{Proposed System}} & \textbf{MERIT} & \textbf{0.130} & \textbf{0.612} & \textbf{0.392} & \textbf{0.298} & \textbf{4.100} & \textbf{3.561} \\
\midrule
\multicolumn{8}{c}{\textbf{Backbone: GLM-5.1}} \\
\midrule
\multirow{11}{*}{\textit{\textbf{Baselines}}}
 & Vanilla LLM                                & 0.099 & 0.609 & 0.379 & 0.113 & 3.523 & 3.644 \\
 & Sliding Window                             & 0.095 & 0.603 & 0.372 & 0.109 & 3.510 & 3.991 \\
 & Conversation Memory                        & 0.097 & 0.608 & 0.375 & 0.110 & 3.781 & 3.934 \\
 \cmidrule{2-8}
 & Session-level RAG                          & 0.100 & 0.610 & 0.378 & 0.111 & 3.818 & 4.157 \\
 & Turn-level RAG                             & 0.106 & 0.613 & 0.386 & 0.115 & 3.712 & 4.195 \\
 \cmidrule{2-8}
 & MemoryBank~\cite{zhong2024memorybank}      & 0.058 & 0.551 & 0.303 & 0.089 & 2.680 & 2.883 \\
 & MemPal~\cite{huang2026mem}                 & 0.071 & 0.582 & 0.374 & 0.094 & 3.365 & 3.481 \\
 & ConditionMem~\cite{yuan2025personalized}   & 0.088 & 0.613 & 0.366 & 0.105 & 4.320 & 3.443 \\
 & RecurSum~\cite{wang2025recursively}        & 0.104 & 0.599 & 0.392 & 0.126 & 4.676 & 4.388 \\
 \cmidrule{2-8}
 & Reflexion~\cite{shinn2023reflexion}        & 0.073 & 0.593 & 0.339 & 0.186 & 3.416 & 3.514 \\
 & ExpeL~\cite{zhao2024expel}                 & 0.079 & 0.601 & 0.341 & 0.211 & 3.314 & 2.958 \\
\midrule
\rowcolor{lightgreen}
\textit{\textbf{Proposed System}} & \textbf{MERIT} & \textbf{0.148} & \textbf{0.638} & \textbf{0.433} & \textbf{0.303} & \textbf{4.739} & \textbf{4.749} \\
\bottomrule
\end{tabular}
\end{table*}

\subsection{Experimental Setup} 
\label{sec:experiments}
\textbf{Implementation Details.} All experiments are conducted on a single NVIDIA A100 80GB GPU. Local models (the policy backbone and the visual extractor) are run on-device, while the GLM-5.1 \cite{zeng2026glm} backbone is accessed through the API. The policy model generates candidate rollouts per turn under stochastic decoding with temperature $0.9$, top-$p$ $0.95$, and a 300-token generation cap; the sampling temperature is incremented by $0.05$ per rollout index to encourage strategic and framing diversity across candidates. Referee model (GPT-5) is accessed through the OpenAI API, queried with temperature $0.2$ for deterministic scoring and an 800-token output cap, with up to three retries under exponential backoff. 

\begin{table}[t]
\caption{Ablation study of \textbf{\textit{MERIT}} with the \textit{Qwen 2.5 3B Instruct} backbone. Each row removes one memory component or the experience rationale library (ExpLib).}
\label{tab:ablation}
\centering
\footnotesize
\setlength{\tabcolsep}{3pt}
\resizebox{\columnwidth}{!}{%
\begin{tabular}{l cccccc}
\toprule
\textbf{Variant} & \textbf{BLEU-2}$\uparrow$ & \textbf{BERTScore}$\uparrow$ & \textbf{ROUGE-1}$\uparrow$ & \textbf{METEOR}$\uparrow$ & \textbf{PAS}$\uparrow$ & \textbf{CAS}$\uparrow$ \\
\midrule
\textbf{MERIT (full)} & \textbf{0.130} & \textbf{0.612} & \textbf{0.392} & \textbf{0.298} & \textbf{4.100} & \textbf{3.561} \\
\midrule
w/o Latent      & 0.064 & 0.560 & 0.299 & 0.191 & 2.559 & 1.966 \\
w/o Strategy    & 0.062 & 0.559 & 0.297 & 0.190 & 2.572 & 2.023 \\
w/o Concession  & 0.064 & 0.561 & 0.301 & 0.193 & 2.540 & 2.076 \\
w/o Session     & 0.054 & 0.548 & 0.282 & 0.171 & 2.244 & 2.185 \\
w/o ExpLib      & 0.062 & 0.560 & 0.297 & 0.189 & 2.559 & 1.990 \\
\bottomrule
\end{tabular}%
}
\end{table}

\textbf{Baselines.} We compare MERIT against memory-augmented dialogue baselines (Memory Bank \cite{zhong2024memorybank}, ConditionMem \cite{yuan2025personalized}, Recursum \cite{wang2025recursively}, and Mem-Pal \cite{huang2026mem} and experience-based reasoning methods (Reflexion \cite{shinn2023reflexion} and ExpeL \cite{zhao2024expel}), along with simple context baselines including a vanilla LLM, a sliding-window model, conversation memory, and session and turn-level retrieval. All methods are evaluated under identical conditions.

\textbf{Evaluation.} We evaluate generated responses using four reference-based metrics: BLEU-2 \cite{papineni2002bleu}, which measures bigram precision; ROUGE-1 \cite{lin2004rouge}, which measures unigram recall; METEOR \cite{banerjee2005meteor}, which captures lexical alignment through exact, stem, and synonym matches; and BERTScore \cite{zhang2019bertscore}, which measures semantic similarity using contextual embeddings. Higher scores indicate greater reference similarity, although these metrics do not directly measure persuasive effectiveness. We further use Gemini 2.5 Pro \cite{comanici2025gemini} as an LLM judge at temperature 0.2 to assign 1--5 scores for Persuasive Appeal Score (PAS), measuring how effectively a response advances the persuasive goal, and Context Alignment Score (CAS), measuring how well it addresses the user's request while respecting implicit needs, persona constraints, and negotiation boundaries. 
Finally, three annotators conduct human evaluation on six parameters. Goal Advancement (GA) measures progress toward the purchase goal, and Long-Term Consistency (LTC) assesses consistency with prior interactions; both are evaluated pairwise against the strongest baselines as Win, Tie, or Loss, with the final label decided by majority vote. Strategic Grounding (SG) assesses the use of identifiable persuasion strategies, Persona Consistency (PC) measures alignment with the persona profile, Engagement (Eng) evaluates conversational involvement, and Fluency (Flu) assesses grammaticality, coherence, and naturalness. The latter four parameters are rated on a 1–5 Likert scale and reported as the mean over annotators and samples, with higher scores indicating better quality.

\begin{table}[t]
\caption{Fine-grained qualitative evaluation on Qwen 3B, rated on a 1--5 scale (higher is better); best result per column in \textbf{bold}. \textbf{SG}: Strategic Grounding; \textbf{PC}: Persona Consistency; \textbf{Eng}: Engagement; \textbf{Flu}: Fluency.}
\centering
\small
\renewcommand{\arraystretch}{1.1}
\begin{tabular}{lcccc}
\toprule
\textbf{Method} & \textbf{SG} & \textbf{PC} & \textbf{Eng} & \textbf{Flu} \\
\midrule
Vanilla LLM & 2.4 & 2.3 & 2.7 & 3.9 \\
MemoryBank       & 3.6 & 3.5 & 3.6 & 4.1 \\
RecurSum    & 3.0 & 2.9 & 3.1 & 4.0 \\
ExpeL       & 3.2 & 3.1 & 3.3 & 4.0 \\

\midrule
\textbf{MERIT} & \textbf{4.3} & \textbf{4.2} & \textbf{4.1} & \textbf{4.3} \\
\bottomrule
\end{tabular}
\label{tab:finegrained}
\end{table}

\begin{table}[t]
\caption{Pairwise comparison of MERIT against baseline systems (Qwen 3B); each cell reports MERIT's \textbf{Win\,/\,Tie\,/\,Loss} counts out of 75 head-to-head dialogues. \textbf{GA}: Goal Advancement; \textbf{LTC}: Long-Term Consistency.}
\centering
\small
\renewcommand{\arraystretch}{1.1}
\begin{tabular}{lcc}
\toprule
\textbf{MERIT vs.} & \textbf{GA} & \textbf{LTC} \\
\midrule
MemoryBank    & 44\,/\,11\,/\,20 & 47\,/\,10\,/\,18 \\
RecurSum & 51\,/\,9\,/\,15  & 54\,/\,8\,/\,13  \\
\bottomrule
\end{tabular}
\label{tab:winrate}
\end{table}

\section{Results}
We summarize our experiments through the following research questions.\par

\begin{figure}[!t]
\centering
\includegraphics[width=\columnwidth]{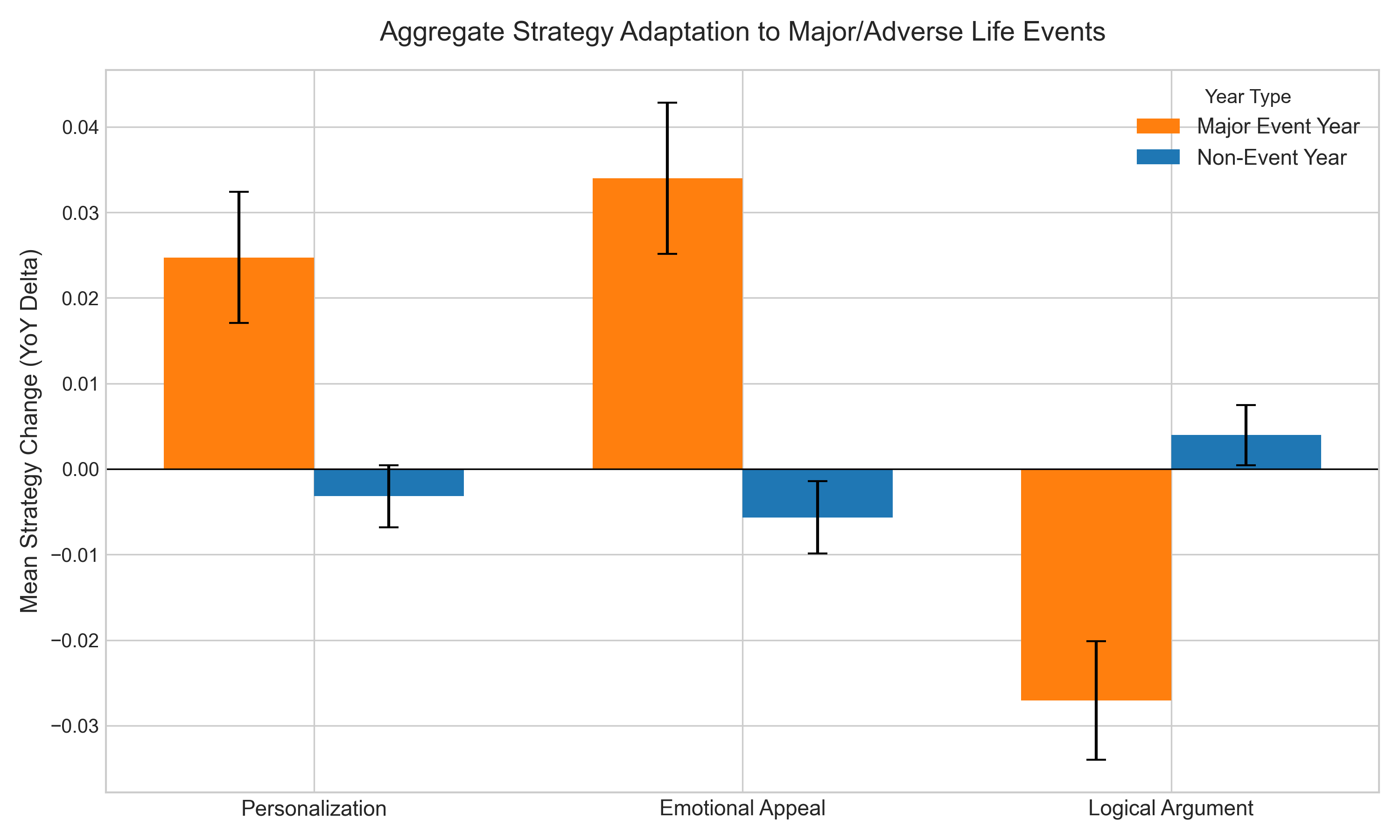}\\[4pt]
\includegraphics[width=\columnwidth]{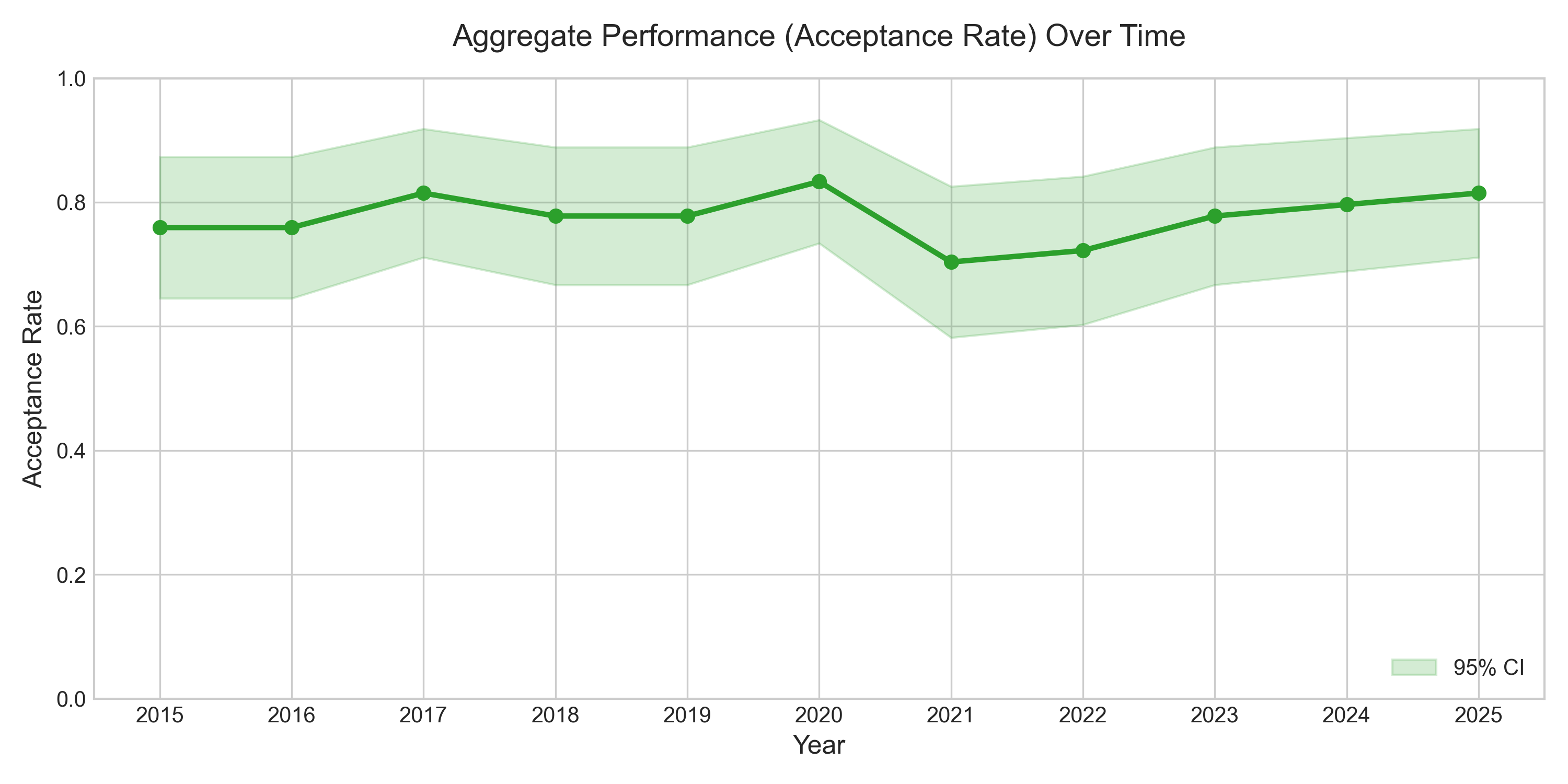}
\caption{Longitudinal dynamics of persuasion in MERIT. 
Top: Mean year-over-year changes in strategy scores during 
major/adverse life-event and non-event years, with 95\% 
confidence intervals. Bottom: Annual acceptance rates from 
2015 to 2025, with 95\% confidence intervals.}
\label{fig:longitudinal}
\end{figure}

\begin{figure}[t]
    \centering
    \includegraphics[width=\columnwidth]{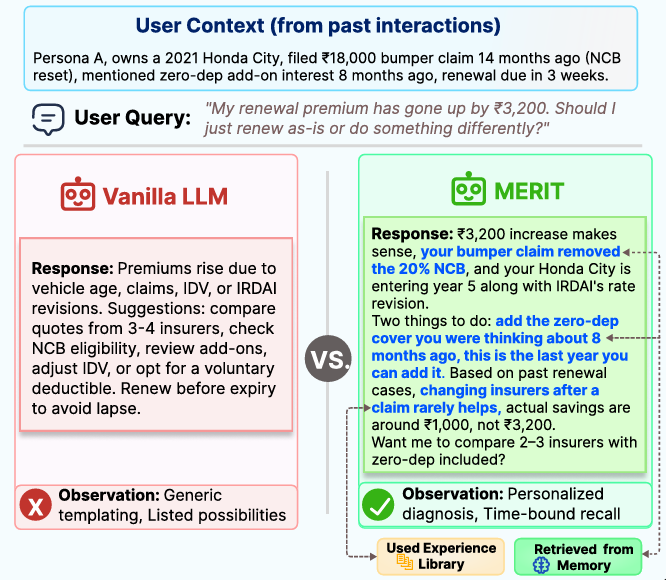}
    \caption{Qualitative comparison on a renewal turn for Persona~A: the vanilla LLM gives a templated reply, while \textbf{\textit{MERIT}} produces a memory-grounded, time-bound, persona-specific diagnosis.}
    \label{fig:qualitative}
\end{figure}

\noindent\textbf{RQ1: MERIT outperforms memory and experience-based baselines.}
On Qwen 2.5 3B (Table~\ref{tab:main_results}), \textbf{\textit{MERIT}} achieves the highest scores across all six metrics (BLEU-2 0.130, BERTScore 0.612, ROUGE-1 0.392, METEOR 0.298, PAS 4.100, and CAS 3.561). While individual baselines perform well on particular metrics, MERIT achieves the best overall results. The findings are consistent on GLM-5.1, where MERIT achieves the highest scores across all six metrics, including PAS 4.739 and CAS 4.749 (Table~\ref{tab:main_results}). Human evaluation further supports these results, with MERIT receiving the highest ratings for Strategic Grounding (4.3), Persona Consistency (4.2), Engagement (4.1), and Fluency (4.3) (Table~\ref{tab:finegrained}). The qualitative example in Figure~\ref{fig:qualitative} shows how MERIT tailors responses to the persona’s history.

\noindent\textbf{RQ2: MERIT's advantage is largest on long-horizon consistency.} Against RecurSum, \textbf{\textit{MERIT}} wins 51/9/15 on Goal Advancement and 54/8/13 on Long-Term Consistency out of 75; against MemoryBank, 44/11/20 and 47/10/18 (Table~\ref{tab:winrate}). LTC win counts exceed GA win counts in both pairings, highlighting the importance of historical consistency. Persona Consistency further confirms this advantage (4.2 vs. RecurSum's 2.9; Table~\ref{tab:finegrained}). Beyond baseline comparisons, major/adverse life-event years show increased Personalization (+0.025) and Emotional Appeal (+0.034), but decreased Logical Argument ($-0.027$) in Figure~\ref{fig:longitudinal}. Meanwhile, annual acceptance rates remain between 70.4\% and 83.3\% over 2015--2025 (Figure~\ref{fig:longitudinal}).

\noindent\textbf{RQ3: The experience library matters as much as memory.} The experience rationale library is as essential as memory itself. Removing it (w/o ExpLib) reduces the judged score from 4.100 to 2.559 on Qwen, a drop comparable to removing any single memory component (2.540--2.572) (Table~\ref{tab:ablation}). Further, \textbf{\textit{MERIT}} outperforms the two experience-based baselines by a wide margin (Reflexion 2.828, ExpeL 2.794 vs. MERIT 4.100 in Table~\ref{tab:main_results}), showing that per-persona rationales are more effective than episode-level reflection. These results demonstrate the importance of incorporating persona-specific persuasive experience into dialogue generation.

\noindent\textbf{RQ4: MERIT's gains hold across model scales.} 
\textbf{\textit{MERIT}} achieves the highest scores across all metrics
on both Qwen-3B and GLM-5.1 (Table~\ref{tab:main_results}).
Its absolute scores also improve with the stronger backbone, with
PAS increasing from 4.100 to 4.739, BLEU-2 from 0.130 to 0.148,
BERTScore from 0.612 to 0.638, and ROUGE-1 from 0.392 to 0.433.
These results show that MERIT's gains hold across different model
families and scales, while a stronger policy further improves
performance. Thus, MERIT remains effective even with a smaller
policy model.


\section{Ablation Study}

\noindent\textbf{Impact of Memory Components.}
Removing any memory type degrades \textbf{\textit{MERIT}} to
near-baseline performance. The session summary is most critical, with
its removal lowering the judge score to 2.244 and BERTScore to 0.548 as
the model loses the episodic record of each prior conversation.
Removing strategy-outcome, latent, or concession memory produces
comparable drops (2.572, 2.559, and 2.540). The closeness of these
values shows that no single memory type dominates: strategy
effectiveness, latent receptiveness, and negotiation behavior each
capture a distinct, non-substitutable facet of persuasion that the
others cannot compensate for.

\noindent\textbf{Impact of the Experience Rationale Library.}
Removing the experience rationale library lowers the judge score to
2.559, a drop comparable to removing any single memory type:
accumulated strategic experience is as essential as memory itself.
\textbf{\textit{MERIT}} also outperforms the experience-based baselines
Reflexion (2.828) and ExpeL (2.794), confirming that its per-persona
rationale mechanism captures strategic knowledge more effectively than
episode-level reflection.
Crucially, these drops are uniform: no component can be removed without
collapsing performance to near-baseline, showing that \textbf{\textit{MERIT's}}
memory and rationales are strongly interdependent.

\section{Conclusion}
We introduced \textbf{\textit{PERSIST}}, a multimodal persona-driven dataset of longitudinal persuasive dialogues, capturing recurring agent-user interactions for 60 personas over 10+ years, with strict human-in-the-loop verification. We proposed \textbf{\textit{MERIT}}, which combines a graph-based memory of four persuasion-specific memory types with a per-persona
experience rationale library capturing which strategies work for a user and why, and adapts a frozen policy model through prompt composition alone. Across automatic and human evaluations, MERIT consistently outperforms memory-augmented and experience-based baselines, and ablations show its components are strongly interdependent. Together, PERSIST and MERIT establish a foundation for studying persuasion as a long-horizon, persona-specific process.

\section{Limitations}


\appendix
\section*{Probabilistic Life Events}
\label{sec:lifeevents}

Each persona is assigned a probabilistic life-event timeline, with events modifying negotiation-relevant attributes. Their effects accumulate over time. Table~\ref{tab:lifeevents} summarizes the events and their corresponding attribute modifications.

\begin{table}[H]
\caption{Probabilistic life events in PERSIST and their corresponding persona-attribute modifications.}
\centering
\small
\begin{tabular}{p{2.3cm}p{4.2cm}}
\toprule
\textbf{Life Event} & \textbf{Attribute Modification (Impact Rule)} \\
\midrule
First car & Sets initial-purchase context; no prior policy \\
Accident \newline (minor / major) & Raises risk appetite toward comprehensive cover; adds protective add-on preference \\
Marriage / Child & Sets priority to family safety; coverage preference to comprehensive \\
Vehicle upgrade & Updates insured value and coverage scope \\
Relocation & Updates risk zone / premium basis \\
Business use & Sets usage type to commercial; raises risk profile \\
Job promotion & Lowers price sensitivity; raises effective budget \\
Job loss & Raises price sensitivity; lowers effective budget \\
Positive claim & Raises trust disposition \\
Negative claim & Lowers trust disposition \\
\bottomrule
\end{tabular}
\label{tab:lifeevents}
\end{table}

\section*{Inter-Annotator Agreement}
\label{sec:IAA}
Inter-annotator agreement for dataset validation was measured using Fleiss' $\kappa$, yielding $\kappa = 0.78$ for life-event plausibility, $\kappa = 0.79$ for chronological consistency, and $\kappa = 0.75$ for consistency between previous interactions and subsequent insurance decisions. For dialogue-quality ratings, ordinal Krippendorff's $\alpha$ yielded $\alpha = 0.80$ for persona consistency, $\alpha = 0.79$ for temporal consistency, $\alpha = 0.81$ for intent alignment, and  $\alpha = 0.77$ for persuasion-strategy appropriateness. For the human evaluation of system outputs, ordinal Krippendorff's $\alpha$ across the four Likert parameters ranged from $\alpha = 0.74$ to $\alpha = 0.79$, and Fleiss' $\kappa$ for GA and LTC was $\kappa = 0.75$ and $\kappa = 0.80$, respectively.

\bibliographystyle{IEEEtran}
\bibliography{references}

\newpage

\vfill

\end{document}